\documentclass[times,10pt,twocolumn]{article}

\usepackage{latex8}
\usepackage{times}
\usepackage{booktabs} 
\usepackage{float}
\usepackage{amsmath}
\usepackage{epsfig}
\usepackage[capitalize]{cleveref}
\DeclareMathOperator*{\argmax}{arg\,max}
\usepackage{subcaption}
\usepackage{caption}
\usepackage{multirow}

\begin{document}

\title{Feature-level and Model-level Audiovisual Fusion for Emotion Recognition \\in the Wild}

\author{Jie Cai$^1$, Zibo Meng$^2$, Ahmed Shehab Khan$^1$, Zhiyuan Li$^1$, James O'Reilly$^1$,\\ Shizhong Han$^3$, Ping Liu$^4$, Min Chen$^5$ and Yan Tong$^1$\\
\small{$^1$Department of Computer Science \& Engineering, University of South Carolina, Columbia, SC}\\
\small{$^2$Innopeak Technology Inc., $^3$12 Sigma Technologies, $^4$JD.com, Inc., CA}\\
\small{$^5$University of Washington Bothell, Bothell, WA}\\
{\footnotesize{\{jcai,akhan,zhiyuanl,oreillyj\}@email.sc.edu, \{mzbo1986,hanshizhong1105,pino.pingliu\}@gmail.com,}}\\
{{\footnotesize{minchen2@uw.edu}, \footnotesize{tongy@cec.sc.edu}}}
}

\maketitle
\thispagestyle{empty}

\begin{abstract}
Emotion recognition plays an important role in human-computer interaction (HCI) and has been extensively studied for decades. Although tremendous improvements have been achieved for posed expressions, recognizing human emotions in ``close-to-real-world'' environments remains a challenge. In this paper, we proposed two strategies to fuse information extracted from different modalities, i.e., audio and visual. 
Specifically, we utilized LBP-TOP, an ensemble of CNNs, and a bi-directional LSTM (BLSTM) to extract features from the visual channel, and the OpenSmile toolkit to extract features from the audio channel.
Two kinds of fusion methods, i,e., feature-level fusion and model-level fusion, were developed to utilize the information extracted from the two channels. 
Experimental results on the EmotiW2018 AFEW dataset have shown that the proposed fusion methods outperform the baseline methods significantly and achieve better or at least comparable performance compared with the state-of-the-art methods, where the model-level fusion performs better when one of the channels totally fails. 
\end{abstract}

\section{INTRODUCTION}
\begin{figure*}[!ht]
\centering
\includegraphics[width=1.4\columnwidth]{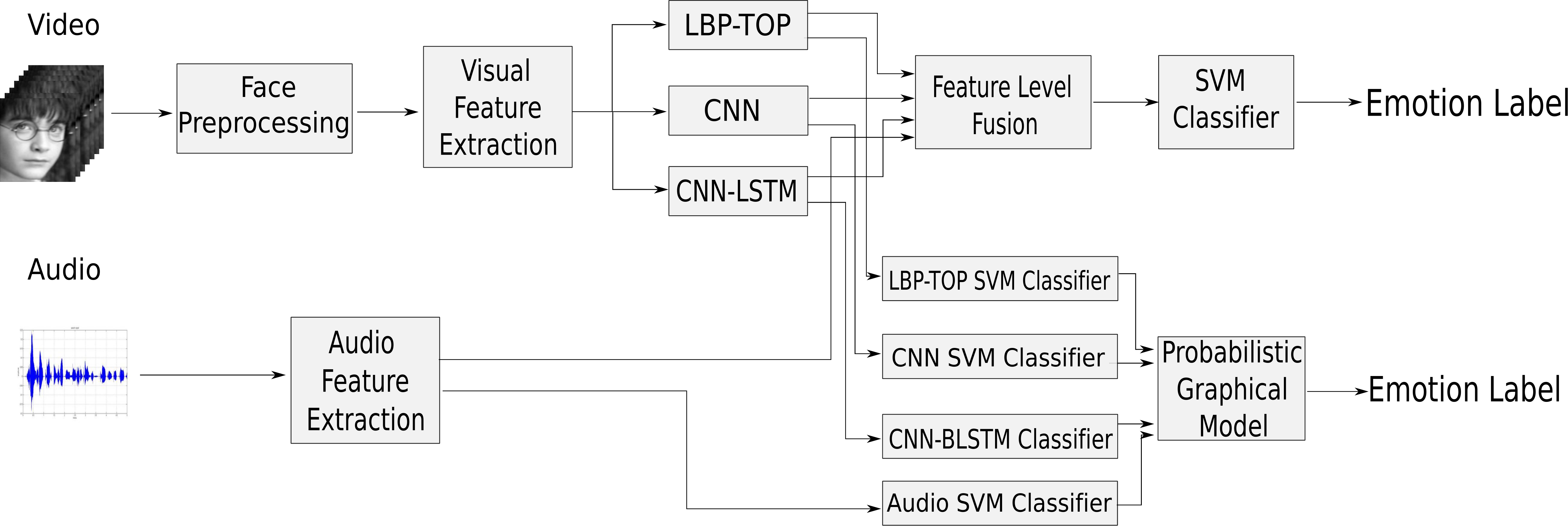}
\caption{The flowchart of the proposed feature-level fusion and model-level fusion framework.}
\label{Fig:audioVisual}
\end{figure*}

Extensive efforts have been devoted to emotion recognition because of a wide range of applications in HCI. Recognizing human emotions from facial displays is still a challenging problem due to head pose variations, illumination changes, and identity-related attributes, which, however, do not affect the information extracted from the audio channel. Human affective behavior is multimodal naturally. Researchers have been investing significant efforts in exploring how to build the best model to effectively exploit information from multiple modalities~\cite{Martinez2017Automatic}. 

We are motivated by the fact that voice and facial expression are highly correlated when conveying emotions. For example, without looking at a face, we would know that a person is happy when hearing laughter. In this paper, we developed a feature-level fusion method and a model-level fusion technique by exploiting the information from both audio and visual channels. For the feature-level fusion, the audio features and three different types of visual features are concatenated to construct a joint feature vector. The model-level fusion explicitly handles differences in time scale, time shift, and metrics from different signals. The audio information and visual information are extracted and employed for emotion recognition independently; and then these unimodal recognition results are combined via a probabilistic framework, i.e., a Bayesian network in this work. Fig.~\ref{Fig:audioVisual} illustrates the proposed audiovisual feature-level fusion and model-level fusion frameworks.

Experimental results on the EmotiW 2018 AFEW database have shown that both of the proposed fusion methods significantly outperform the baseline unimodal recognition methods and also achieve better or comparable performance compared with the state-of-the-art methods.

\section{RELATED WORK}

Facial activity plays a significant role in emotion recognition. One of the major steps is to extract the most discriminative features that are capable of capturing appearance and geometry facial changes. These features can be roughly divided into two categories: human designed and learned features. 
Most recently, CNNs~\cite{han2018optimizing,meng2017identity,cai2018probabilistic,cai2018island,cai2019identity,xue2017differential,xue2018deep,li2019pooling} have achieved promising recognition performance under real-world conditions as demonstrated in the recent EmotiW2017~\cite{knyazev2017convolutional,hu2017learning,vielzeuf2017temporal} and EmotiW2018~\cite{khan2018group} challenges.

Speech/voice is another important means for human communication. 
As elaborated in the survey paper~\cite{Zeng2009}, the most popular features for audio-based emotion recognition include prosodic features, e.g., pitch-related features, energy-related features, zero crossing, formant, teager energy operator, fundamental frequency (F0), speech rate, and spectral features. Studies show that pitch-related and energy-related features are the most important audio features for affect recognition. Most recently, some spectrum features, such as Linear Prediction Coefficients~\cite{makhoul1975linear}, Linear Prediction Cepstrum Coefficients~\cite{atal1974effectiveness}, Mel-Frequency Cepstrum Coefficients~\cite{davis1980comparison} and its first derivative, are employed for emotion recognition. RASTA-PLP~\cite{hermansky1991rasta} is another popular audio feature, which combines Relative Spectral Transform~\cite{hermansky1994rasta} and Perceptual Linear Prediction~\cite{hermansky1990perceptual}.

Furthermore, audiovisual fusion for emotion recognition has received an increasing interest. Previous approaches on audiovisual emotion recognition can be roughly divided into three categories: feature-level fusion, decision-level fusion, and model-level fusion.

\emph{Feature-level fusion} directly combines audio and visual features into a joint feature vector~\cite{Zeng2009} and trains a classifier on top of it for emotion recognition. Recently, deep learning has been utilized for learning features from both audio and visual input~\cite{kaya2015contrasting,ebrahimi2015recurrent}. For example, Kahou et al.~\cite{ebrahimi2015recurrent} employed audio features, aggregated CNN features, and RNN features as input for a Multilayer Perceptron (MLP). Yao et al.~\cite{yao2015capturing} utilized the AU-aware features and their latent relations and then fed the concatenated features into a group of SVMs to make prediction.


\emph{Decision-level fusion} combines recognition results from two modalities assuming that audio and visual signals are conditionally independent of each other~\cite{Zeng2009,yao2015capturing,ebrahimi2015recurrent,fan2016video, yan2016multi}, while there are strong semantic and dynamic relationships between audio and visual channels. For example, Yao et al.~\cite{yao2015capturing,yao2016holonet} performed the decision-level fusion by averaging prediction scores of individual classification scores. In~\cite{kaya2015contrasting,ebrahimi2015recurrent,fan2016video,yan2016multi,hu2017learning}, fusion by weighted scores of individual models were used instead of taking average over all different models.

\emph{Model-level fusion} exploits correlation between audio and visual channels~\cite{Zeng2009} and is usually performed in a probabilistic manner. For example, coupled, tripled, or multistream fused HMMs~\cite{Zeng2008} were developed by integrating multiple component HMMs, each of which corresponds to one modality, e.g., audio or visual, respectively. Fragpanagos et al.~\cite{Fragopanagos2005} and Caridakis et al.~\cite{Caridakis2006} used an ANN to perform fusion of different modalities. Sebe et al.~\cite{Sebe2006} employed a Bayesian network to recognize expressions from audio and facial activities. Chen et al.~\cite{chen2014emotion} employed Multiple Kernel Learning (MKL) to find an optimal combination of the features from two modalities.

\section{METHODOLOGY}

\subsection{Audiovisual Feature Extraction}

In our work, information is extracted from both audio and visual channels. 
Specifically, for the audio channel, we utilized a set of low-level spectral or voicing related feature descriptors and voice/unvoice durational features; for the visual channel, we employed both human-crafted features, i.e., LBP-TOP~\cite{Zhao2007}, and features learned by a CNN and a bi-directional LSTM. Finally, we combined audio and visual information for emotion recognition given a video clip.

\subsubsection{Audio Features}

In this work, audio signal was extracted from video at a sampling rate of 48 kHz and 160k bps. Then, a feature set of 1582 audio features, which was used in the INTERSPEECH 2010 Paralinguistic challenge~\cite{schuller2011avec}, was extracted by OpenSMILE~\cite{eyben2013recent}. The feature set is comprised of 34 spectral related low-level audio feature descriptors (LLDs) with their delta coefficients $\times21$ functionals, 4 voicing related LLDs with their delta coefficients $\times19$ functionals, and 2 voiced/unvoiced durational features. In addition, PCA was utilized to compress the features. Specifically, the first 20 principal components were preserved.


\subsubsection{LBP-TOP-based Visual Features}

Following the EmotiW 2018 baseline paper~\cite{dhall2018EmotiW2018}, we extracted LBP-TOP features~\cite{Zhao2007} ($3\times 59$ dimensions) from non-overlapping spatial $4\times 4$ blocks. The LBP-TOP features from all blocks are concatenated to create one feature vector. 
Then, we employed PCA for dimensionality reduction and preserved the first 150 principal components.

\subsubsection{Visual Features by An Ensemble of CNNs}

\begin{figure}[!ht]
\centering
\includegraphics[width=0.75\columnwidth]{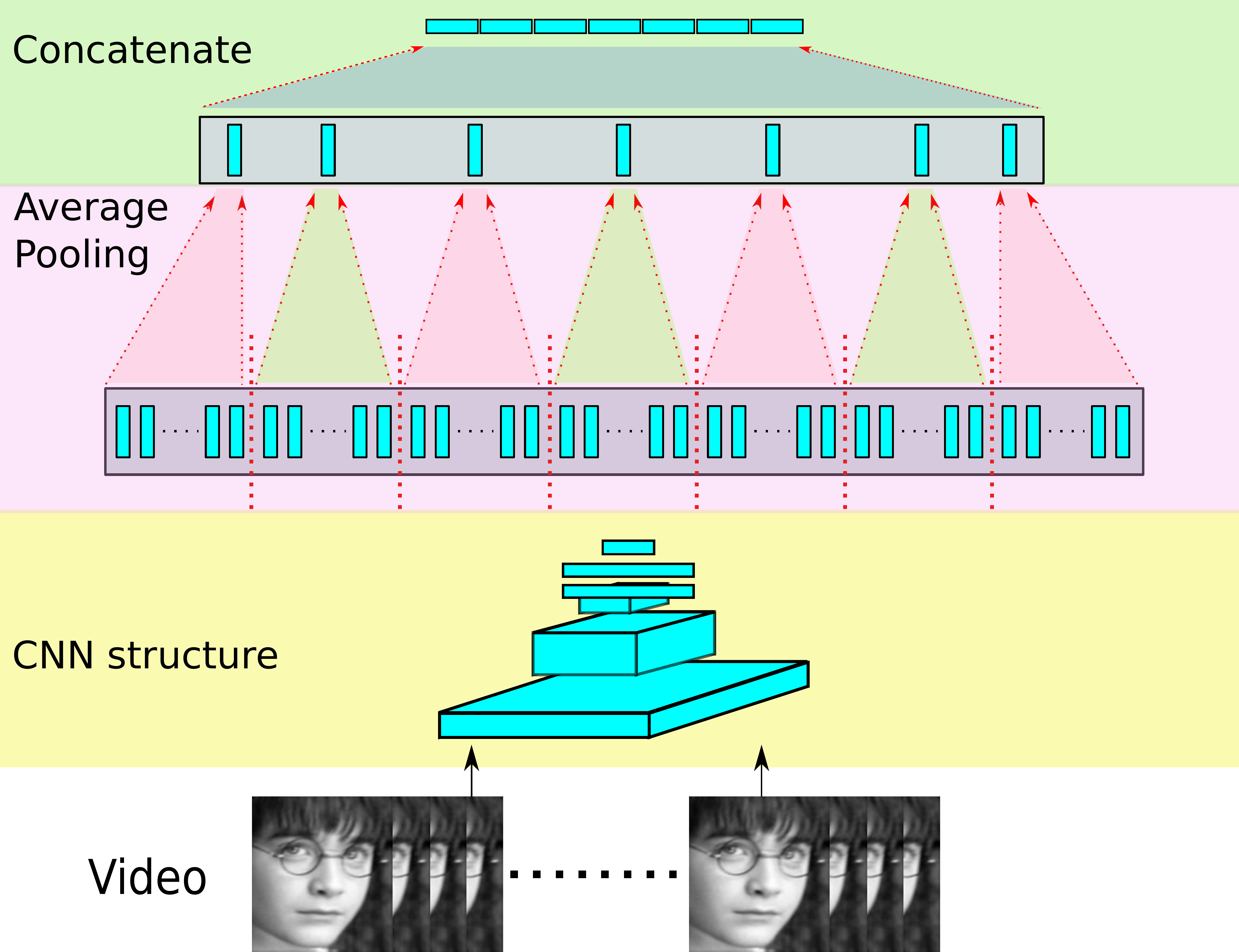}
\caption{An illustration of $k$-average temporal pooling.}
\label{Fig:CNNfusion}
\end{figure}

In this work, two CNN architectures, i.e., a VGG-Face network~\cite{parkhi2015deep} and a shallow CNN~\cite{cai2018island}, were employed as our backbone CNNs. Starting from the pre-trained VGG-Face model~\cite{parkhi2015deep}, the VGG-Face CNN was first fine-tuned on the CK+ dataset~\cite{Lucey2010}, the MMI dataset~\cite{pantic2005web}, the Oulu-CASIA dataset~\cite{Zhao2011Facial}, the RAF-DB dataset~\cite{Li2017reliable}, and ExpW dataset~\cite{zhang2018facial}, and then fine-tuned on the AFEW training set. The shallow CNN was pre-trained on the aforementioned five datasets and also the FER-2013 dataset~\cite{goodfellow2013challenges}, and then fine-tuned on the AFEW training set.

To further improve performance, an island loss~\cite{cai2018island}, which was designed to simultaneously reduce the intra-class variations and increase the inter-class differences, and the softmax loss were jointly used as the supervision signal to train the CNNs. The island loss denoted as $\mathcal{L}_{IL}$ is defined as the summation of the center loss~\cite{wen2016discriminative} and the pairwise distances between class centers in the feature space:

\begin{footnotesize}
\begin{equation} \label{eq:island_loss_forward}
\mathcal{L}_{IL} =  \frac{1}{2}\sum\limits_{i=1}^{m} \| \textbf{x} _{i} - \textbf{c} _{y_{i}} \|^{2}_{2} + \lambda_{1}  \sum\limits_{\textbf{c}_{j} \in \mathcal{N}} \sum\limits_{ \substack{\textbf{c}_{k} \in \mathcal{N} \\ ~\textbf{c}_{k} \neq \textbf{c}_{j} }} \left(\frac{\textbf{c}_{k} \cdot \textbf{c}_{j}}{ \|\textbf{c}_{k}\|_{_2} \|\textbf{c}_{j}\|_{_2} } +1 \right)
\end{equation}
\end{footnotesize}
where $\mathcal{N}$ is the set of emotion labels; $\textbf{c}_{k}$ and $\textbf{c}_{j}$ denote the $k^{th}$ and $j^{th}$ emotion center with $L_2$ norm $\|\textbf{c}_{k}\|_{_2}$ and $\|\textbf{c}_{j}\|_{_2}$, respectively; $(\cdot)$ represents the dot product. Specifically, the first term penalizes the distance between the sample and its corresponding center, and the second term penalizes the similarity between emotions. $\lambda_1$ is used to balance the two terms. By minimizing the island loss, samples of the same emotion will get closer to each other and those of different emotions will be pushed apart.

Inspired by ensemble learning techniques, we built an ensemble of 40 CNNs, half of which employed the VGG-Face structures and the others used the shallow CNN structures. In addition, for each network structure, half of the models used both the softmax loss and the island loss, while the others used only the softmax loss.
Each frame was fed into the 40 trained CNNs. Then the average scores of the softmax layer outputs of the 40 CNNs are employed as the per-frame feature.

In order to capture temporal information, we employed k-average temporal pooling as described in~\cite{ebrahimi2015recurrent}. The per-frame CNN features are averaged into $k$ bins to generate a k-dimensional vectors as the per-video-clip feature. In this work, $k$ was set to $7$ empirically. As a result, the per-video-clip CNN features are 49 dimensions. For videos with less than $k$ frames, the frames are locally repeated until the image length reaches $k$. In addition, when the number of frames cannot be evenly divided by $k$, several frames at the head and the tail of each video clip are discarded. Fig.~\ref{Fig:CNNfusion} illustrates how the k-average temporal pooling works.

\subsubsection{Visual Features Learned by a CNN-BLSTM}

\begin{figure}[!ht]
\centering
\includegraphics[width=0.8\columnwidth]{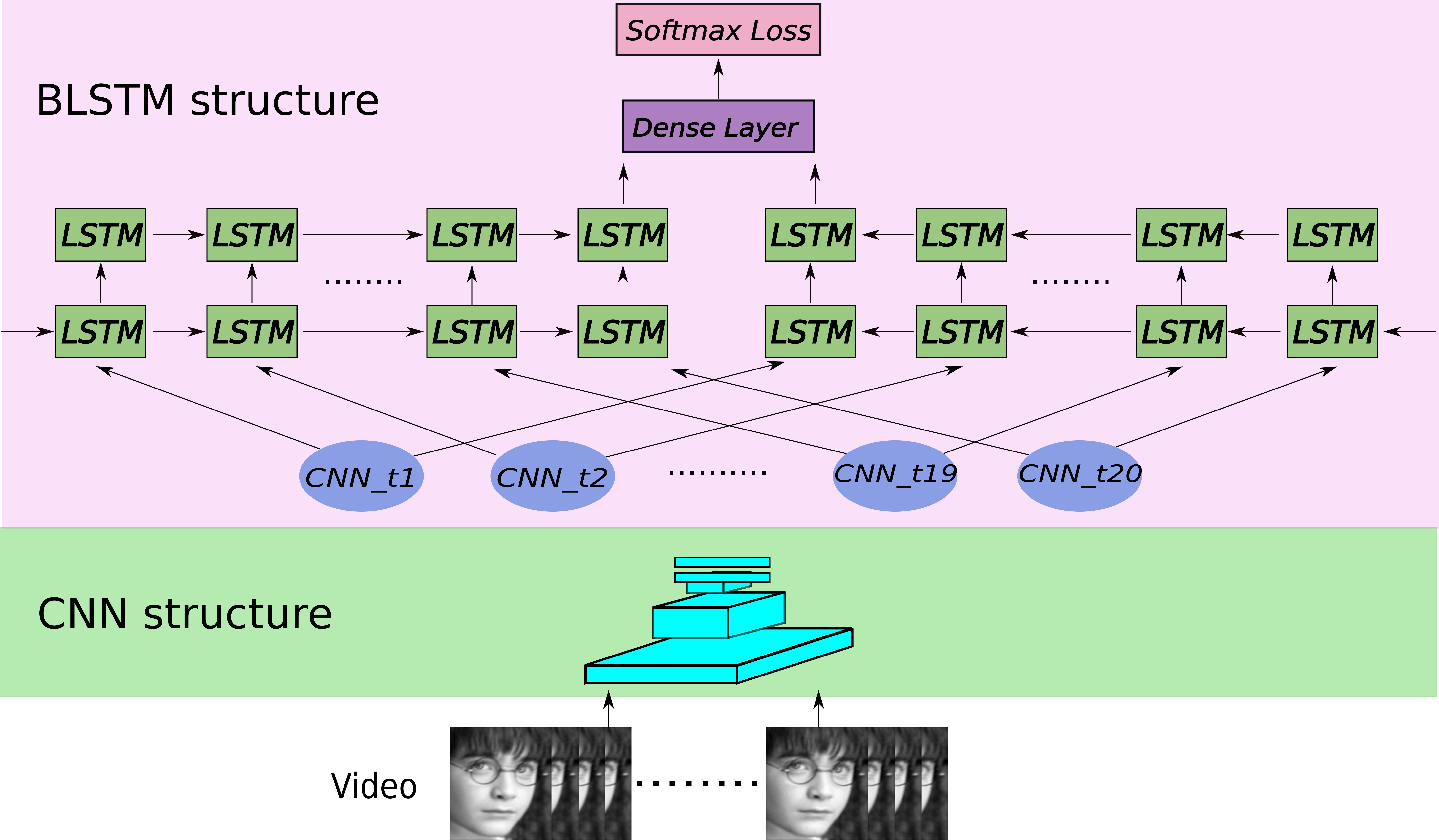}
\caption{The proposed CNN-BLSTM framework.}
\label{Fig:lstm}
\end{figure}

As shown in Fig.~\ref{Fig:lstm}, a hybrid CNN-BLSTM network was designed for video-level prediction. Specifically, CNN features of the fine-tuned VGG-Face model were fed into a two-layer BLSTM. Next, features from both directions of the BLSTM were concatenated and then connected to a dense layer (512 dimensions). Finally, a softmax layer was employed to predict the emotion label of the video sequence. During training, a series of 20 images were randomly chosen from each video clip and their CNN features were used as the BLSTM inputs. For testing, a series of 20 evenly spaced images were chosen for each video clip. For videos with less than 20 frames, the frames are locally repeated until the image length reaches 20.

\subsection{Audiovisual Feature-Level Fusion}

\begin{figure}[!th]
\centering
\includegraphics[width=0.6\columnwidth]{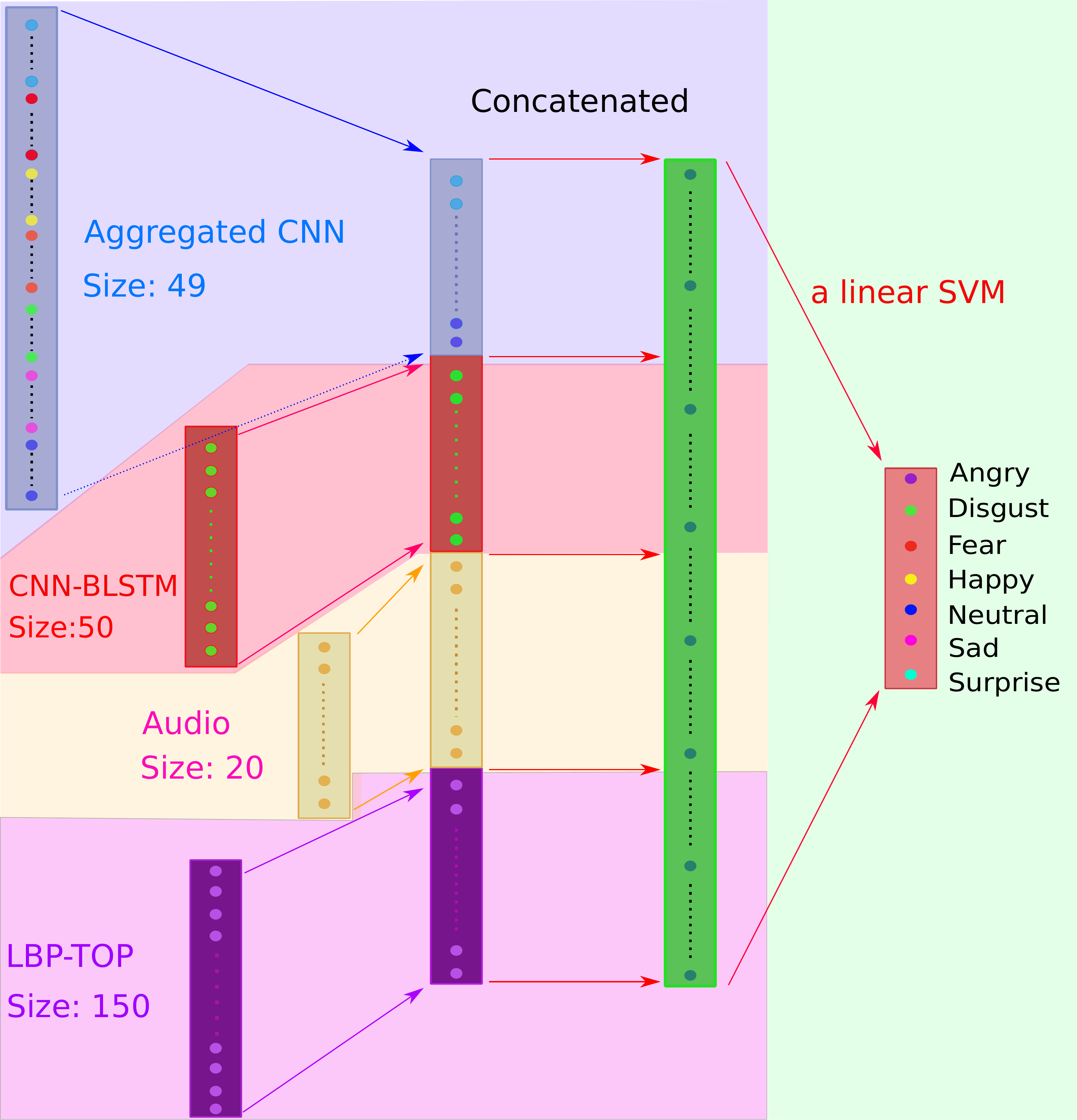}
\caption{The proposed feature-level fusion framework.}
\label{Fig:fusion}
\end{figure}

In our work, the four types of features are directly extracted from the whole video clip and concatenated to form a joint feature vector. As shown in Fig.~\ref{Fig:fusion}, the input features to the feature-level fusion are: (1) the first 20 principal components of the audio features, (2) the first 150 principal components of the LBP-TOP features, (3) the 49-dimensional ($7\times 7$ bins) CNN features, and (4) the first 50 principal components of the BLSTM features. Hence, we can obtain a joint feature vector consisting of 269 features for each video clip, which was fed to a linear SVM for emotion classification.

In order to deal with the difference in metrics when extracting features from the four models/channels, the features were first normalized for each of the 269 dimensions to zero mean and unit variance across all samples. Then, each feature in the joint feature vector was normalized to zero mean and unit variance across all 269 dimensions.

\vspace{-0.1in}
\subsection{Audiovisual Model-Level Fusion}
\vspace{-0.1in}
\begin{figure}[!ht]
\centering
\includegraphics[width=0.6\columnwidth]{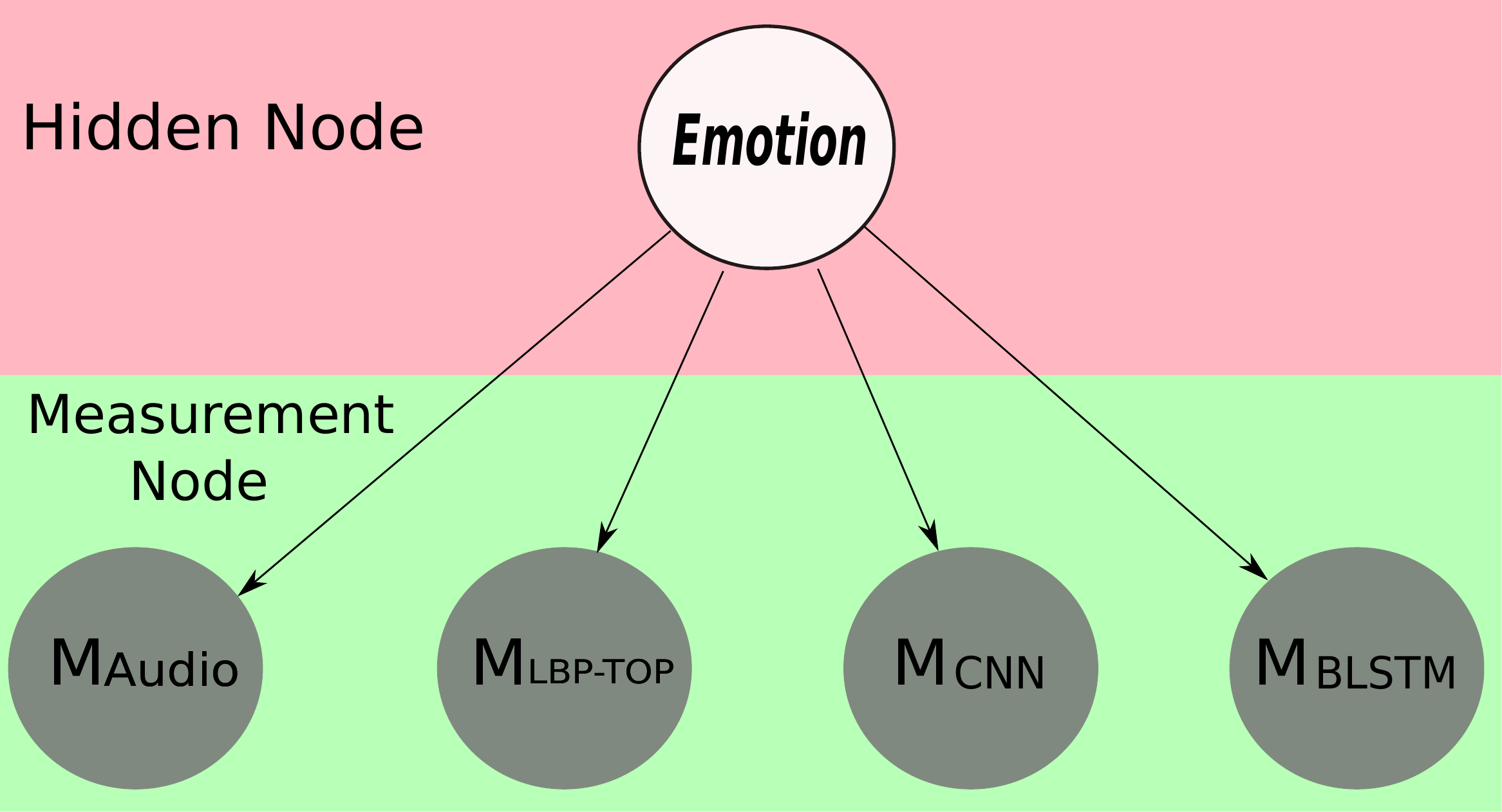}
\caption{A BN model for the model-level fusion. The unshaded node represents the hidden node of emotion. The shaded nodes are measurements from the four classifiers of emotion recognition. The directed links between the hidden node and the measurement nodes represent measurement uncertainty associated with each classifier.}
\label{Fig:BN}
\end{figure}

In this work, we employed a Bayesian Network (BN) for model-level fusion as illustrated in Fig.~\ref{Fig:BN}. A BN is a directed acyclic graph, where the shaded nodes are measurement nodes whose states are available during inference and the unshaded node is the hidden node whose state can be estimated by probabilistic inference over the BN.

Particularly, the four measurement nodes correspond to the four emotion recognition methods employing the four types of features, i.e., audio features, LBP-TOP features, CNN features, and CNN-BLSTM features, respectively. SVMs were employed for classification using each type of features. The directed links between the hidden node ``Emotion'' and the measurement nodes represent the measurement uncertainty associated with each measurement node, i.e., the recognition accuracy using each method, which can be estimated on the validation set. During inference, the final decision is made by maximizing the posterior probability given observations from all measurements:

\begin{scriptsize}
\begin{equation} \label{eq:BN}
{\mathrm{Emotion}}^{*} \!=\! \argmax P\!\left (\!\mathrm{Emotion} |M_{_{ \mathrm{Audio}} }\!,\!M_{_{\mathrm{LBP-TOP}}}\!,\!M_{_{\mathrm{CNN}}}\!,\!M^{}_{_{\mathrm{BLSTM}}}\!\right )
\end{equation}
\end{scriptsize}

\section{EXPERIMENTS}

\subsection{Experimental Dataset}

The proposed fusion methods were evaluated on the Acted Facial Expressions in the Wild (AFEW)~\cite{dhall2012collecting} dataset. The AFEW dataset is a dynamic facial expression dataset consisting of short video clips, which are collected from scenes in movies/TV shows with natural head pose movements, occlusions and various illuminations. Each of the video clips has been labeled as one of seven expressions: anger, disgust, fear, happiness, sadness, surprise, and neutral. The subjects are diverse in race, age, and gender. The dataset is divided into three partitions: a training set (773), a validation set (383), and a test set (653) such that the data in those three sets are coming from mutually exclusive movies and actors. The test set labels are held by the challenge organizers and unknown to the public.

\subsection{Preprocessing}

Face alignment was employed on each image based on centers of two eyes and nose. The aligned facial images were then resized to $60\times 60$ for the shallow CNN and $256\times 256$ for the VGG-Face CNN. For data augmentation purpose, $48\times 48$ patches and $224\times 224$ patches were randomly cropped from the $60\times 60$ images and $256\times 256$ images, and then rotated by a random degree between -5$^\circ$ and 5$^\circ$, respectively. The rotated images were horizontally flipped randomly as the input of all CNNs.

\subsection{Experimental Results}

The proposed two fusion methods were evaluated and compared with the four baseline methods based on (1) audio, (2) LBP-TOP, (3) an ensemble of CNNs, and (4) CNN-BLSTM, respectively. As shown in Table~\ref{tab:experimental_results}, the deep learning based features (the ensemble CNN features and the CNN-BLSTM features) achieved better recognition performance than the human crafted audio/visual features among the baseline methods. Both the proposed fusion methods improve the baseline unimodal methods employing a single type of features with a large margin.

\begin{table}[!ht]
\centering
\caption{Performance comparison on the validation and test sets in terms of the average recognition accuracy of the 7 emotions.}
\label{tab:experimental_results}
 \scalebox{0.9}{
	\begin{tabular}{c|c|c|c} \hline
	Method                   		          &	      Channel      &    Validation       	&     Test     \\ \hline
	Hu et. al~\cite{hu2017learning}	          &	    audio, visual  &     59.01   	        &      \textbf{60.34}           \\ \hline
	Fan et. al~\cite{fan2016video}             &	    audio, visual  &     --  	            &        59.02           \\ \hline
	Vielzeuf et. al~\cite{vielzeuf2017temporal}&	    audio, visual  &     --  	            &        58.81           \\ \hline
	Yao et. al~\cite{yao2016holonet}	          &	    audio, visual  &     51.96   	        &        57.84           \\ \hline
	Ouyang et. al~\cite{ouyang2017audio}	      &	    audio, visual  &     --      	        &        57.20           \\ \hline
	Kim et. al~\cite{kim2017multi}	          &	    audio, visual  &     50.39     	        &        57.12           \\ \hline
    Yan et. al~\cite{yan2016multi}	          &	    audio, visual  &     --     	            &        56.66           \\ \hline
	Wu et. al~\cite{wu2016multi} 	          &	    audio, visual  &     --      	        &        55.31           \\ \hline
	Kaya et. al~\cite{kaya2017video} 	      &	    audio, visual  &     57.02   	        &        54.55           \\ \hline
	Ding et. al~\cite{ding2016audio}	          &	    audio, visual  &     51.20   	        &        53.96           \\ \hline
	Yao et. al~\cite{yao2015capturing}	      &	    audio, visual  &     49.09   	        &        53.80           \\ \hline
	Kaya et. al~\cite{kaya2015contrasting}	  &	    audio, visual  &     52.30   	        &        53.62           \\ \hline
	Kahou et. al~\cite{ebrahimi2015recurrent}  &	    audio, visual  &  	 --                 &        52.88           \\ \hline
	Sun et. al~\cite{sun2016lstm}	          &	    audio, visual  &     --    	            &        51.43           \\ \hline
	Pini et. al~\cite{pini2017modeling}	      &	    audio, visual  &     49.92    	        &        50.39           \\ \hline
	Li et. al~\cite{li2015deep}	              &	    audio, visual  &     --    	            &        50.46           \\ \hline
	Gideon et. al~\cite{gideon2016wild}	      &	    audio, visual  &     38.81    	        &        46.88           \\ \hline
	\hline
	Bargal et. al~\cite{bargal2016emotion}	  &	       visual      &  \textbf{59.42}   	    &        56.66           \\ \hline
	Sun et. al~\cite{sun2016facial}	          &	       visual      &       50.67    	        &        50.14           \\ \hline	
	\hline
	Audio (baseline)	                     &      audio      &      35.51          &          --              \\ \hline
	LBP-TOP (baseline)				     &      visual      &      38.90	        &          --              \\ \hline
	CNN (baseline)	                     &      visual      &      47.00		    &	       --              \\ \hline
	CNN-BLSTM (baseline)                  &     visual      &       49.09		    &	       --              \\ \hline
	\hline
	Feature-level fusion	         		&     audio, visual  &       53.79	        &          \textbf{56.81}         \\ \hline
	Model-level fusion	         		&     audio, visual  &       \textbf{54.83}	&          54.06                  \\ \hline
\end{tabular}
}
\end{table}

\vspace{-0.1in}
\subsection{Analysis of Fusion Methods }
\vspace{-0.1in}
As shown in Table~\ref{tab:experimental_results}, the feature-level fusion method achieved the best result with the performance of 56.81\% on the test set; while the model-level fusion achieved a better result on the validation set compared to the feature-level fusion. This may be because the measurement uncertainty measured from the validation set is different from that on the test set.

Although the average recognition performance was improved significantly by using the fusion-based methods compared to all of the baseline methods, the improvement for recognizing disgust, fear, and surprise was marginal as compared to the best baseline methods, i.e., the two deep learning based methods on the validation set. Furthermore, the feature-level fusion method failed to recognize disgust, fear, and surprise on the test set. This is because all of the four types of features could not well characterize these three expressions: the recognition accuracies of the four baseline methods are all below $50\%$ for these expressions. In contrast, by considering the measurement uncertainty in a probabilistic manner, the model-level fusion yielded the best results on recognizing these difficult expressions on the validation set. These observations imply that the feature-level fusion may further boost the recognition performance of those expressions that can be well recognized by a single type of features, while the model-level fusion may be employed to improve the recognition performance of the difficult expressions.

\section{CONCLUSION}

We proposed two novel audiovisual fusion methods by exploiting audio features, LBP-TOP-based features, CNN-based features, and CNN-BLSTM features. Both the proposed fusion methods significantly outperform the baseline methods that employ a single type of feature in terms of the average recognition performance. In the future, more advanced techniques will be developed to improve the audio-based emotion recognition by exploring deep learning based approaches.


\section{Acknowledgement}

This work is supported by National Science Foundation under CAREER Award IIS-1149787. The Titan Xp used for this research was donated by the NVIDIA Corporation.

{
\fontsize{9.5pt}{10.5pt}
\bibliographystyle{latex8.bst}
\bibliography{../../../../bibliography/abbrev_short,../../../../bibliography/machine_learning/ty-literature_graphical_model,../../../../bibliography/ty-literature_misc,../../../../bibliography/ty-literature_self,../../../../bibliography/emotion/ty-literature_AU_Exp_Emotion_rec,../../../../bibliography/ty-literature_audiovisual_ASR,../../../../bibliography/ty-literature_facial_feature_detect_track,../../../../bibliography/machine_learning/ty-literature_unsupervised_feature_learning,../../../../bibliography/ty-literature_database,../../../../bibliography/machine_learning/ty-literature_machine_learning,../../../../bibliography/machine_learning/ty-literature_deep_learning,../../../../bibliography/machine_learning/ty-literature_metric_learning,../../../../bibliography/ty-literature_statstical_models_alignment,../../../../bibliography/object_classification/ty-literature_object_detection,../../../../bibliography/ty-literature_EmotiW,../../../../bibliography/ty-literature_EmotiW_extra,../../../../bibliography/other/ty-literature_EmotiW2018,../../../../bibliography/other/ty-literature_other_paper}
}

\end{document}